\documentclass{llncs}
\usepackage{graphicx}
\usepackage{url}

\title{ Automated Reasoning and Presentation Support for Formalizing Mathematics
in Mizar\thanks{The final publication of this paper is
  available at www.springerlink.com}}

\author{Josef Urban$^1$\thanks{Supported by the NWO
project ``MathWiki a Web-based Collaborative Authoring Environment for
Formal Proofs''.} \and Geoff Sutcliffe$^2$}
\institute{ Radboud University, Nijmegen
\and University of Miami}
\begin{document}
\maketitle

\begin{abstract}
This paper presents a combination of several automated
reasoning and proof presentation tools with the Mizar system for
formalization of mathematics. The combination forms an online
service called {\sf MizAR}, similar to the
{\sf SystemOnTPTP} service for first-order automated reasoning. 
The main differences to {\sf SystemOnTPTP} are the use of the Mizar language
that is oriented towards human mathematicians (rather than the pure
first-order logic used in {\sf SystemOnTPTP}), and setting the service in
the context of the large Mizar Mathematical Library of previous theorems, 
definitions, and proofs (rather than the isolated problems that are solved
in {\sf SystemOnTPTP}).
These differences poses new challenges and new opportunities for automated 
reasoning and for proof presentation tools. 
This paper describes the overall structure of {\sf MizAR}, 
and presents the automated reasoning systems and proof presentation tools 
that are combined to make {\sf MizAR} a useful mathematical service.
\end{abstract}
\section{Introduction and Motivation}

Formal mathematics, in its interactive and verification aspects, and in
the automated reasoning aspect, is becoming increasingly well-known, used,
and experimented with \cite{Hal08}.
Projects like FlySpeck \cite{Hal05}, formal proof of the Four Color 
Theorem \cite{Gon08}, verification of tiny (but real) operating systems 
\cite{KE+09}, and the increased use of verification for software and
hardware \cite{DKW08}, are stimulating the development of interactive
verification tools and interactive theorem provers (ITPs).
Linked to this is the development of strong automated theorem proving (ATP)
systems, used either independently to solve hard problems in suitable domains
\cite{McC97,PS08,BP10}, or integrated with interactive tools
\cite{Pau99,Hur03,CCK+07}.
ATP development has also stimulated interesting research in the context of
automated reasoning in large theories \cite{PS07-ESARLT,US+08,SS+09}.

The goal of the work presented here is to make formal mathematics and
automated reasoning easily accessible to practitioners in these
areas, by putting most of the work into their browsers, and providing a very
fast (real-time) server-based experience with a number of ATP, ITP,
presentation, and AI tools that work well together.
This is important for supporting existing users and attracting new users 
of Mizar, by providing them with an attractive environment for 
exploring the world of formal reasoning in mathematics.
Fast server-based solutions make systems easy to use, to the extent of
just ``pushing a button'' (clicking on a HTML link), rather than having to
go through the pains of building an adequate local hardware and software
installation, for benefits that might initially not be clear.
Server-based solutions are becoming an important part of general computer 
use, and formal mathematics is no exception.
It is not possible to name all the server-based services that already exist
for informal mathematics, starting e.g., from the arXiv, Wikipedia,
MathOverflow, PolyMath, Wolfram MathWorld, PlanetMath, ProofWiki, the SAGE
system for working with CASes, etc.

This paper describes the {\em Automated Reasoning for Mizar} ({\sf MizAR})
web service, which combines several automated reasoning and proof presentation 
tools with the Mizar system for formalization of mathematics, to form a useful 
mathematical service.
{\sf MizAR} runs in the context of the Mizar Mathematical Library (MML),
and uses the Mizar language that is oriented towards human mathematicians.
The main inspiration for {\sf MizAR} is the {\sf SystemOnTPTP} ATP service
\cite{Sut00-CADE-17}.
{\sf SystemOnTPTP} allows users to easily experiment with many
first-order ATP systems in a common framework, and provides additional services
such as proof presentation with the IDV system \cite{TPS06}, discovery of
interesting lemmas with the AGInT system \cite{PGS06}, and independent
proof verification with the GDV verifier \cite{Sut06}.
Pieces of the {\sf SystemOnTPTP} infrastructure also served in the initial
implementation of the {\sf MizAR} web service.
{\sf SystemOnTPTP}'s infrastructure is briefly described in 
Section~\ref{SystemOnTPTP}.
Section~\ref{Description} describes the implemented {\sf MizAR}
service, and demonstrates its use.
Section~\ref{Future} considers a number of possible future extensions, and
concludes.

\section{{\sf SystemOnTPTP}}
\label{SystemOnTPTP}

The core of {\sf SystemOnTPTP} is a utility that allows an ATP problem or 
solution to be easily and quickly submitted in various ways to a range of 
ATP systems and tools.
{\sf SystemOnTPTP} uses a suite of currently available systems and tools,
whose properties (input format, reporting of result status, etc) are stored
in a simple text database.
The input can be selected from the TPTP (Thousands of Problems for Theorem 
Provers) problem library or the TSTP (Thousands of Solutions from Theorem 
Provers) solution library \cite{Sut09}, or provided in TPTP format 
\cite{SS+06} by the user.
The implementation relies on several subsidiary tools to preprocess the 
input, control the execution of the chosen ATP system(s), and postprocess 
the output.
On the input side {\sf TPTP2X} or {\sf TPTP4X} is used to prepare the input
for processing.
A strict resource limiting program called {\tt TreeLimitedRun} is used to
limit the CPU time and memory used by an ATP system or tool.
{\tt TreeLimitedRun} monitors processes' resource usage more tightly than
is possible with standard operating system calls.
Finally a program called {\tt X2tptp} converts an ATP system's output to TPTP
format, if requested by the user.

The web interfaces {\sf SystemB4TPTP}, {\sf SystemOnTPTP}, and
{\sf SystemOnTSTP} provide interactive online access to the
{\sf SystemOnTPTP} utility.\footnote{%
Available starting at \url{http://www.tptp.org/cgi-bin/SystemOnTPTP}}
The online service can also be accessed directly with {\tt http POST}
requests.
The {\sf SystemB4TPTP} interface provides access to tools for preparing
problems for submission to an ATP system, including conversion from other
(non-TPTP) formats to TPTP format, parsing and syntax checking, type checking,
and pretty printing.
In addition to providing access to ATP systems, the {\sf SystemOnTPTP}
interface additionally provides system reports, recommendations for
systems to use on a given problem, and direct access to the {\sf SSCPA}
system \cite{SS99-FLAIRS} that runs multiple systems in competition parallel.
The {\sf SystemOnTSTP} interface provides access to solution processing tools,
including parsing and syntax checking, pretty printing, derivation
verification using GDV \cite{Sut06}, interactive graphical proof presentation
using IDV \cite{TPS06}, answer extraction \cite{SYT09}, and proof
conversion and summarization tools.
The three interfaces have options to pass the output from a system/tool
on to the next interface -- from problem preparation, to problem solving,
to solution processing.
The output is returned to browsers in appropriate HTML wrapping, and can
also be obtained in its raw form for processing on the client side (typically
when the interfaces are called programmatically using {\tt http POST} requests).
The online service is hosted at the University of Miami on a server with
four 2.33GHz CPUs, 4GB RAM, and running the Linux 2.6 operating system.

\section{{\sf MizAR}}
\label{Description}

{\sf MizAR} is running experimentally on our server\footnote{%
\url{http://mws.cs.ru.nl/~mptp/MizAR.html}},
where it can be best learned by exploration.
A good way to explore is to start with an existing simple Mizar article,
e.g., the {\tt card\_1} article\footnote{%
\url{http://mws.cs.ru.nl/~mptp/mml/mml/card_1.miz}} 
about cardinal numbers \cite{Ban90}\footnote{%
Available at \url{http://mizar.uwb.edu.pl/JFM/Vol1/ordinal1.html}}, 
from the MML.\footnote{%
All articles are available from \url{http://mws.cs.ru.nl/~mptp/mml/mml}}
Within {\sf MizAR}, select the ``URL to fetch article from'' field,
insert the article's URL into the text box, and press the ``Send'' button.
For experienced Mizar users, there is also a simple way to send the current
Mizar buffer to the remote service, by running the
\texttt{mizar-post-to-ar4mizar} interactive function in the Mizar mode for
Emacs \cite{Urb06-JAL}.
Both actions call the main {\tt MizAR} {\tt cgi-bin} script with appropriate 
arguments, which launches the functions described below.

{\sf MizAR} links together a number of Mizar and ATP-related components,
which are useful for general work with formal mathematics in Mizar.
The main components are as follows (details are provided in the rest of
Section~\ref{Description}):
\begin{itemize}
\item Web access to the whole cross-linked HTMLized MML.
\item Fast server-based verification of a Mizar article.
\item Disambiguation of the article by HTMLization, producing an HTML
      presentation of the verified article with links to an HTMLized 
      version of the MML. 
      Additional useful information is also extracted during HTMLization, and
      included in the HTML presentation of the article.
\item Fast translation of the article to MPTP (Mizar Problems for Theorem
      Provers) format.
\item Fast generation of ATP problems in TPTP format, for all the theorems 
      in the article, and for all the atomic inferences done by Mizar.
\item Easy access to default ATP systems for solving the ATP problems,
      and access to {\sf SystemOnTPTP} for solving more difficult problems.
\item Easy access to IDV for visualization and postprocessing of proofs 
      found by the ATP systems.
\item Suggesting useful hints for proving (either by ATP or interactively in
      Mizar) particular Mizar lemmas and theorems.
\end{itemize}

\begin{figure}[tb!]
\begin{center}
    \includegraphics[width=\textwidth]{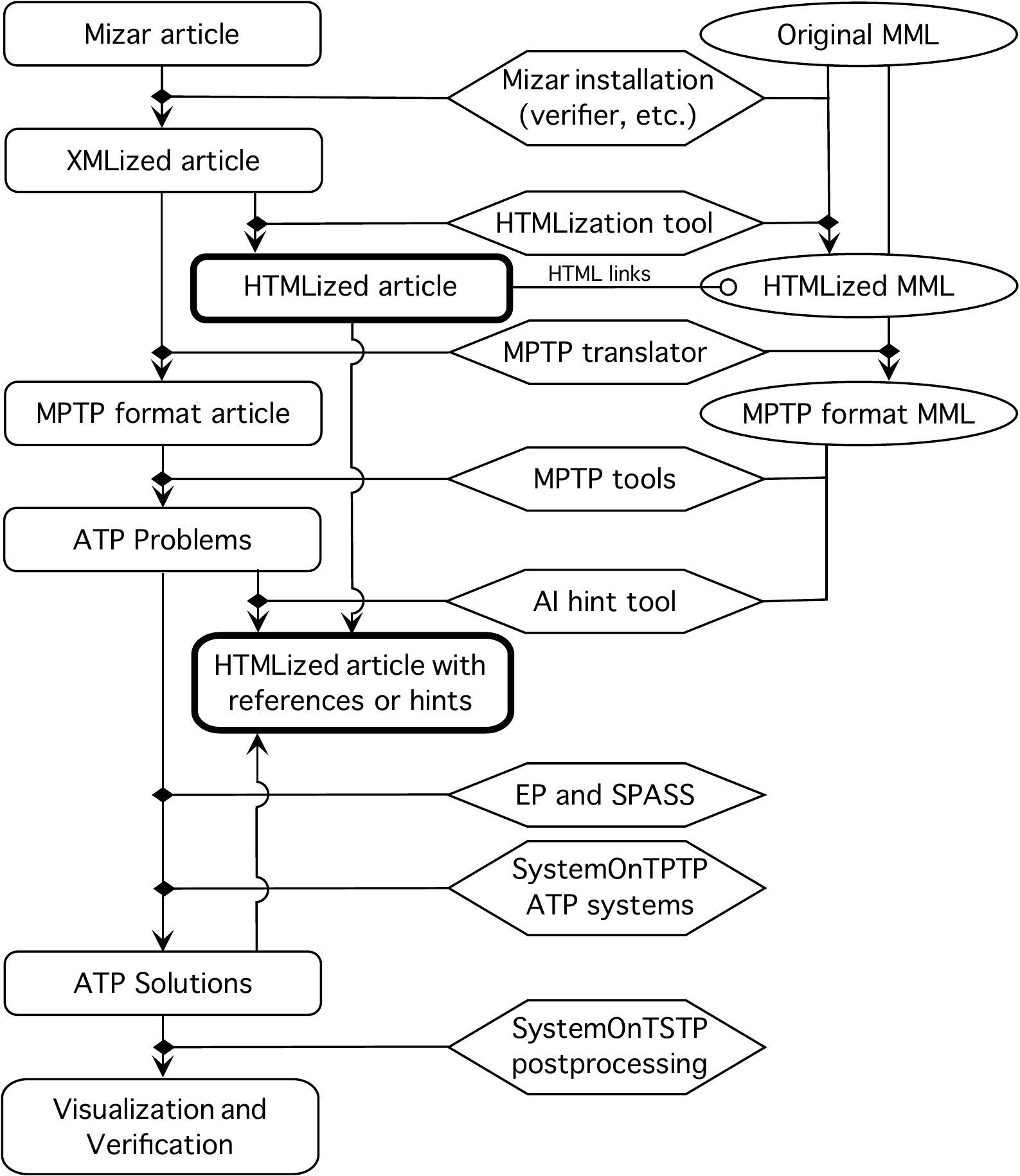}
  \caption{Structure of the {\sf MizAR} system}
  \label{Structure}
\end{center}
\end{figure}

Figure~\ref{Structure} shows the overall structure of the {\sf MizAR} system. 
The leftmost column shows the various forms of the article that are
produced, and the two bold boxes in the next column are the HTML presentations
for user interaction.
The third column shows the software tools that generate the various dataforms,
using the article and the background information shown in the rightmost column.
A Mizar article is submitted through the web interface or from Emacs.
The article is then verified and converted to XML format, which is subsequently
rendered in HTML format with links to the MML.
The HTML presentation includes links that allow the user to proceed with 
further processing, and is the main interface for user interaction with
{\sf MizAR}.
While the HTML is presented to the user, the article is asynchronously
converted to the MPTP format, which is used for generating TPTP format
ATP problems.
The ATP problems can then be submitted to ATP systems, either locally
via {\sf SystemOnTPTP}.
The ATP systems' solutions are used to enrich the HTML presentation, and
can be passed on to various post-processing tools.
The subcomponents that perform these tasks are described in more detail below.

\subsection{Server-based Verification of a Mizar Article}

Unlike many other (especially LCF-inspired) proof assistants, Mizar is a
compiler-like batch processor, verifying a whole article in one
pass.
While a lot of work on Mizar goes into balancing the strength, speed,
and obviousness of the proof checking, the process of checking a whole
article can get quite time-consuming for longer and more complex articles,
especially on older hardware.

There are several advantages to remote server-based verification of
Mizar articles.
The first obvious advantage is that having everything web-based removes
the need for a local installation of Mizar.
The second advantage is that even if Mizar is installed locally, it is
often more convenient to quickly (try to) verify an article in a browser,
instead of launching the verification environment on one's local computer.
In cases when the article is available online, it is possible to provide
that URL as an argument to the {\sf MizAR} URL, to directly launch {\sf MizAR}
on the article.\footnote{%
For example, \url{http://mws.cs.ru.nl/~mptp/cgi-bin/MizAR.cgi?ProblemSource=URL&FormulaURL=http://mws.cs.ru.nl/~mptp/mml/mml/card_1.miz&Name=Test1}}
This makes such verification snippets available in all kinds of online fora
(e-mail discussions, blog and twitter posts, wikis, etc.) with direct
rendering of the results.
In short, the third advantage is that a web service provides mechanism
for online communication of verification results.

The fourth (and probably greatest) advantage of server-based verification
is the raw verification speed.
A dedicated server usually runs on reasonably new hardware with enough
memory, etc.
For example, even for the relatively short {\tt card\_1} Mizar article
mentioned above, full verification on a recent notebook (1.66GHz Intel
Atom) takes 2s, while on a recent lower-range server (2 quad-core
hyperthreading 2.27GHz Intel Xeons) the same task takes 0.5s.
For a more involved article this difference becomes more visible, and can
be the deciding factor for the usability and real-time experience with
the system.
For example, for the more involved Mizar article {\tt fdiff\_1}\footnote{%
\url{http://mws.cs.ru.nl/~mptp/mml/mml/fdiff_1.miz}} about real function 
differentiability \cite{RS90}, the difference is 23s vs. 6s.

The latest advances in CPU power have been achieved mainly by packing 
multiple CPUs together, instead of raising the speed of individual CPUs.
To take advantage of this, Mizar processing has recently been
parallelized\footnote{%
The description of the Mizar parallelization and related experiments is
unpublished as of January 2010.
The parallelizer is available at
\url{http://github.com/JUrban/MPTP2/raw/master/MizAR/cgi-bin/bin/mizp.pl}.},
and the parallel version of the Mizar verifier is running on our server.
This (depending on the success of the parallelization) can further
significantly improve the real-time verification experience.
For example, on the even longer Mizar article {\tt fdiff\_2}\footnote{%
\url{http://mws.cs.ru.nl/~mptp/mml/mml/fdiff_2.miz}} about real function 
differentiability \cite{KR91}, the difference between running the parallel 
version (using eight cores) and the non-parallel version is a factor of 
four (31s vs. 7.8s).
Verification of this article using the above mentioned notebook takes 125s,
resulting in a speed-up factor of sixteen (and substantially improving the
real-time interaction with the system).

The last important advantage is that a server-based installation
supports use of modified, enhanced, and experimental versions of the verifier.
This can provide useful additional functionalities.
For instance, the Mizar parallelizer requires additional
software to run, and a recently modified version of the Mizar verifier
that has not yet been distributed to Mizar users.
Translation of Mizar articles to ATP formats also requires a version of
the verifier that has been compiled with a special flag, again not included
in the standard Mizar distribution.
An online service can also easily include multiple versions of the Mizar 
library and binaries, as is done for the MML Query service \cite{BR03}.

\subsection{HTMLization of Mizar Articles}
\label{HTML}

There has been quite a lot of recent work on XMLization and HTMLization
of Mizar articles \cite{Urb05-MKM,US+09}, including the addition
of useful additional information into the XML form of the Mizar article
and its HTML presentation.
There are two major reasons for having a static HTMLized MML 
available:\footnote{%
It is available at \url{http://mws.cs.ru.nl/~mptp/mml/html/}}
(i)~to provide fast browsing of the theorems and
definitions used in a particular formalization, with a number of user-friendly
features (like (sub)proof hiding/showing, etc.), and
(ii)~providing explanations for a number of phenomena in the formalization
that are made explicit and clear only during verification, and are hard
to decipher from the formalization text alone.
The latter includes, for example:
\begin{itemize}
\item Explicit HTML presentation of the current goal (thesis), computed by
      the verifier at each point of the formalization.
\item Proper disambiguation of overloaded mathematical symbols. 
      Overloading is necessary in a large body of mathematics including
      all kinds of subfields, but at the same time makes it difficult for 
      readers of the textual versions of the articles to understand the 
      precise meaning of the overloaded symbols.
\item Explicit access to formulae for definition correctness, and 
      formulae expressing properties (projectivity, antisymmetry, etc.) 
      that are computed by the verifier.
      Making these explicit in the HTML presentation can help users.
\item Explicit representation of other features that are implicit
      in Mizar verification, e.g., definitional expansions, original
      versions of constructors that have been redefined, etc.
      Making these explicit in the HTML presentation can also help users.
\end{itemize}

The static HTMLized MML is an important resource used by {\sf MizAR}.
The articles submitted to {\sf MizAR} are dynamically linked to the
static HTMLized MML.
This is a notable difference to {\sf SystemOnTPTP}, which treats
each problem as an independent entity.
The first implementation of {\sf MizAR} has focused on
developing the services for a fixed version of the MML.
However, management of library versions is a nontrivial and interesting
problem. 
Allowing users to verify new articles and also refactor
existing ones will ultimately lead into the area of formal mathematical
wikis \cite{CK07,KC+08}.

The main functions of the HTMLization service are to (i)~provide
quick linking to the static HTMLized MML (thus providing the
disambiguation and explanation functions described above), and (ii)~allow
a number of additional (mainly automated reasoning and AI) services to be
launched by suitable CGI and AJAX calls from links in the HTML.
These additional services are described in the following subsections.
The main features of server-side HTMLization are increased speed, and 
the availability of additional programs and features.
While the Mizar HTML processing was designed to be locally available,
using just a browser-based XSL processor (i.e., loading the XML produced
by Mizar directly into a browser, which applies the appropriate style sheet),
even the basic XSL processing in the browser can take a long time (minutes).
Again, having a specialized fast XSL processor installed on the server helps
quite a lot, and the HTMLization can be parallelized using techniques similar
to the parallelization of the basic verification process.
This provides a much better HTMLization response, and also makes additional
XSL-based preprocessing possible.
This is needed for better HTML quality, and for the translation to ATP
formats.

\subsection{Generation of ATP Problems in TPTP Format}
\label{TPTP}

One of the main objectives of {\sf MizAR} (as suggested by its name) 
is to allow easy experimentation with ATP systems over the large body of
formal mathematics available in the MML, and to apply ATP functionalities 
on Mizar articles.
The MPTP system~\cite{Urb08-KEAPPA,Urb06} for translating Mizar articles
to the TPTP format has been modified to work in a fast real-time mode,
generating ATP problems corresponding to Mizar proof obligations.
The MPTP system translates Mizar articles to the MPTP format, which is 
an extension of the TPTP format with information needed for further
processing into ATP problems.
Like the static HTMLized MML, a static copy of the MML in MPTP format is
available to {\sf MizAR}.
It is used for building translated MML items (theorems, definitions,
formulae encoding Mizar type automations, etc.) that are necessary for
creating complete ATP problems.

Using MPTP and generating ATP problems requires a quite complex installation 
and setup (SWI Prolog, Unix, special XSL style sheets, the translated MML
in the MPTP format, etc.), so this is a good example of an additional
functionality that would be quite hard to provide locally.
The MPTP was initially designed for offline production of interesting
ATP problems and data, and was not optimized for speed.
Several techniques have been used to provide a reasonable real-time
experience:
\begin{itemize}
\item More advanced (graph-like) data structures have been used to
      speed up the selection of parts of the MML necessary for generating 
      the ATP problems.
\item Larger use has been made of Prolog indexing and the asserted database,
      for various critical parts of the code.
\item The MPTP version of the MML has been factored so that it is possible
      to work with only the parts of the MML needed for a given article.
\end{itemize}

These techniques have led to reasonable real-time performance of the MPTP
problem generation, comparable to the performance of Mizar verification
and HTMLization.
For example, the MPTP processing followed by the generation of all 586 
ATP problems for the {\tt card\_1} article takes 7s on the server.

After the conversion to MPTP format, the ATP problems for an article are 
generated asynchronously while the user is presented with the HTMLized 
article.
There is a small danger of the user wanting to solve an ATP problem that
has not yet been generated.
However, it is easy to check if the translation process is finished, and
which ATP problems are already available, by examining the system log.

The MPTP processing has not been parallelized (like the Mizar
verification and HTMLization).
However, there are no serious obstacles to that.
Another speed-up option would be to ask MPTP to generate only a subset of
the ATP problems (this is actually how the parallelization is going to work),
selected in some reasonable way by the formalizer in the user interface.
It is also possible to keep the MPTP system loaded and listening once
it has generated a subset of problems, and to have it generate new ATP 
problems from the current article on demand.

The HTMLization of an article and the generation of ATP problems are 
independent processes that could be separated into two separate services.
Users might, for instance, be interested only in HTML-like disambiguation
of their articles, or only in getting explanations and advice from
ATP systems, without looking at the HTML form of the article.
With sufficient CPU-cores in the server, none of these two possible use-cases
suffers in terms of the response time.

\subsection{Calling ATP Systems}
\label{ATPCalls}

The calling of ATP systems to solve the ATP problems is built into the
HTML presentation of the user's article, by linking the available ATP 
services to keywords in the HTML presentation.
This follows the general idea that the HTML serves as the main interface
for calling other services.
The links to the ATP services in the HTML are the Mizar keywords \textbf{by}
and \textbf{from}, indicating semantic justification in Mizar.
For example, the Mizar justification
\begin{verbatim}
    thus ( f is one-to-one & dom f = X & rng f = A )
       by A1, A4, WELLORD2:25, WELLORD2:def 1;
\end{verbatim}
in the last line of the proof of theorem {\tt Th4} in the {\tt card\_1} article
says that the Mizar checker should be able to verify that the formula
on the left hand side of the \textbf{by} keyword follows from the previously
stated local propositions, theorems and definitions \texttt{A1, A4,
WELLORD2:25, WELLORD2:def 1}, and some knowledge that the Mizar verifier
uses implicitly.
There are now the following use-cases calling ATP systems:
\begin{enumerate}
\item Mizar has verified the inference, possibly using some implicit
      information.
      The user is interested in knowing exactly what implicit information
      was used by Mizar, and exactly how the proof was conducted.
\item Mizar has not verified the inference.
      The user is interested in knowing if the inference is logically valid,
      if it can be proved by a (stronger) ATP system, and what such an ATP
      proof looks like.
\end{enumerate}

\begin{figure}[bt!]
\begin{center}
    \includegraphics[width=\textwidth]{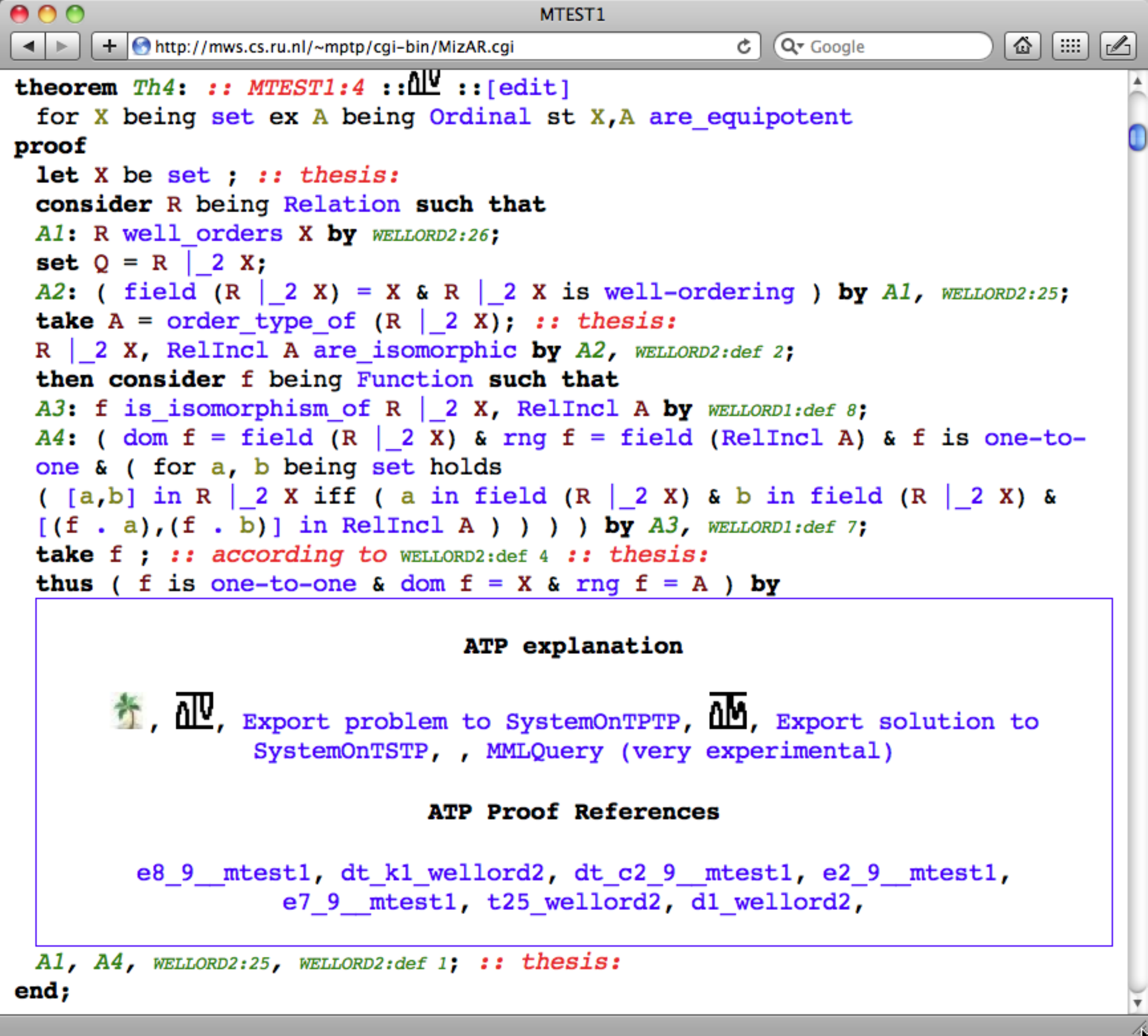}
  \caption{ATP explanation box}
  \label{ByExpl}
\end{center}
\end{figure}

The first use-case typically happens for one of two reasons.
The first reason is that the theory in which the author is working has
become very rich, and involves many implicit (typically typing) Mizar
mechanisms that make the formal text hard to understand.
The TPTP translation has to make all this implicit information explicit
in the TPTP problems, and the resulting corresponding ATP proofs show
explicitly how this information is used.
For the Mizar justification above, clicking on the \textbf{by} keyword
calls the EP system~\cite{Sch02-AICOMM} on the ATP problem, with a several
second time limit.
If a proof is found, the interface is refreshed with an explanation box that
includes (among other things described below) a list of the references
used in the proof, as shown in Figure~\ref{ByExpl}.
In this case the exact references shown to the user are following:
\begin{verbatim}
    e8_9__mtest1, dt_k1_wellord2, dt_c2_9__mtest1, e2_9__mtest1,
    e7_9__mtest1, t25_wellord2, d1_wellord2
\end{verbatim}
These references use the MPTP syntax, but are linked (dynamically using
AJAX calls) to the corresponding places in the theorem's HTML or the
static HTMLized MML), and are given appropriate explanation titles.
Note that the EP proof uses seven more references than the four that are 
in the original Mizar \textbf{by} inference.
One reference is added because it explicitly denotes the formula being
proved (the left-hand side of \textbf{by}), and the two remaining references
encode implicit type declarations that are used by Mizar (the type of the
local constant \texttt{R}, and the type of the functor \texttt{RelIncl} that
is used in proposition \texttt{A4} (renamed to \texttt{e7\_9\_\_mtest1}
by MPTP)).
The ATP proof can be visualized in the IDV system by clicking on the palm tree
icon in the explanation box.

The second reason for the first use-case is cross-verification.
In cases when a bug in the Mizar implementation is suspected, or
incompleteness in the ATP translation is suspected, the user may be
interested in knowing if the Mizar proof can be done by another system
(and how).
In this sense the environment is used for gathering additional information
and debugging.
The cross-verification rates for Mizar justifications are reasonably high
\cite{US09}, which makes this usage realistic.

\begin{figure}[b!]
\begin{center}
    \includegraphics[width=\textwidth]{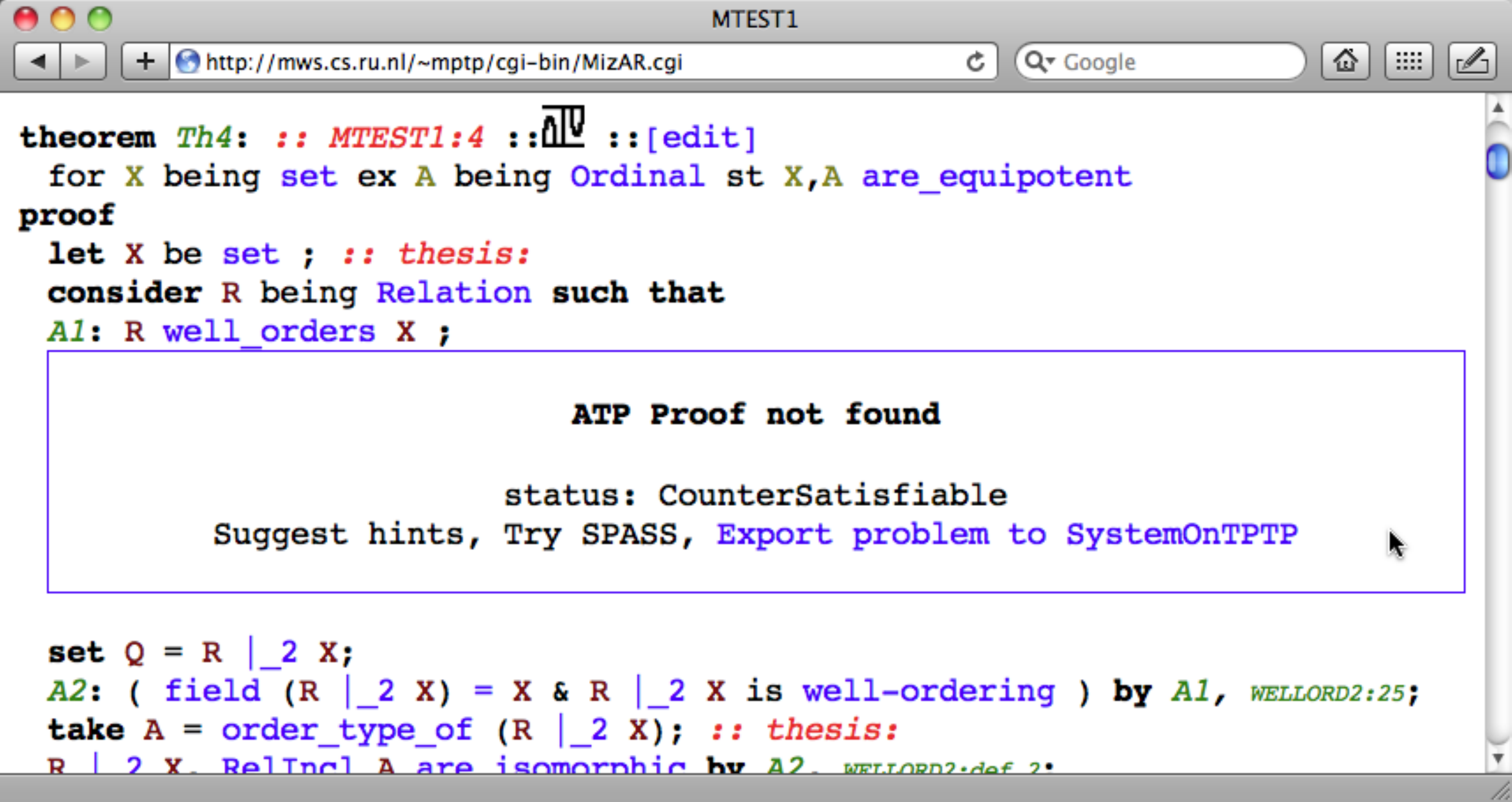}
  \caption{ATP explanation box for ``Proof not found''}
  \label{SPASS}
\end{center}
\end{figure}

The second use-case (finding proofs that are too hard for Mizar) is the
real ``ATP proof assistance'' dream, i.e., using ATP systems to automatically
find proofs for ITPs.
Users can do this within {\sf MizAR} by providing a large set of
``potentially relevant'' Mizar propositions on the right-hand side of
the \textbf{by} keyword, and letting the EP system try to find a proof.
Note that if EP does not find the problem to be countersatisfiable, the
user also has the option to try the SPASS ATP system \cite{WS+07} directly 
from the interface, as shown in Figure~\ref{SPASS}.
This is justified by the general experience that SPASS is reasonably
complementary to EP when solving MPTP problems.
If SPASS is not successful the user can use the links and icons in the
explanation box to inspect the ATP problem, and launch the
{\sf SystemOnTPTP} interface to try the ATP systems available there.
The proofs found by the ATP systems can be processed in the {\sf SystemOnTSTP}
interface, including visualization using the IDV system, analysis using
the AGInT system for finding the interesting steps in proofs, and
ATP-based cross-verification using the GDV verifier.

\subsection{Getting Hints for Necessary Mizar References}
\label{GettingHints}

\begin{figure}[b!]
\begin{center}
    \includegraphics[width=\textwidth]{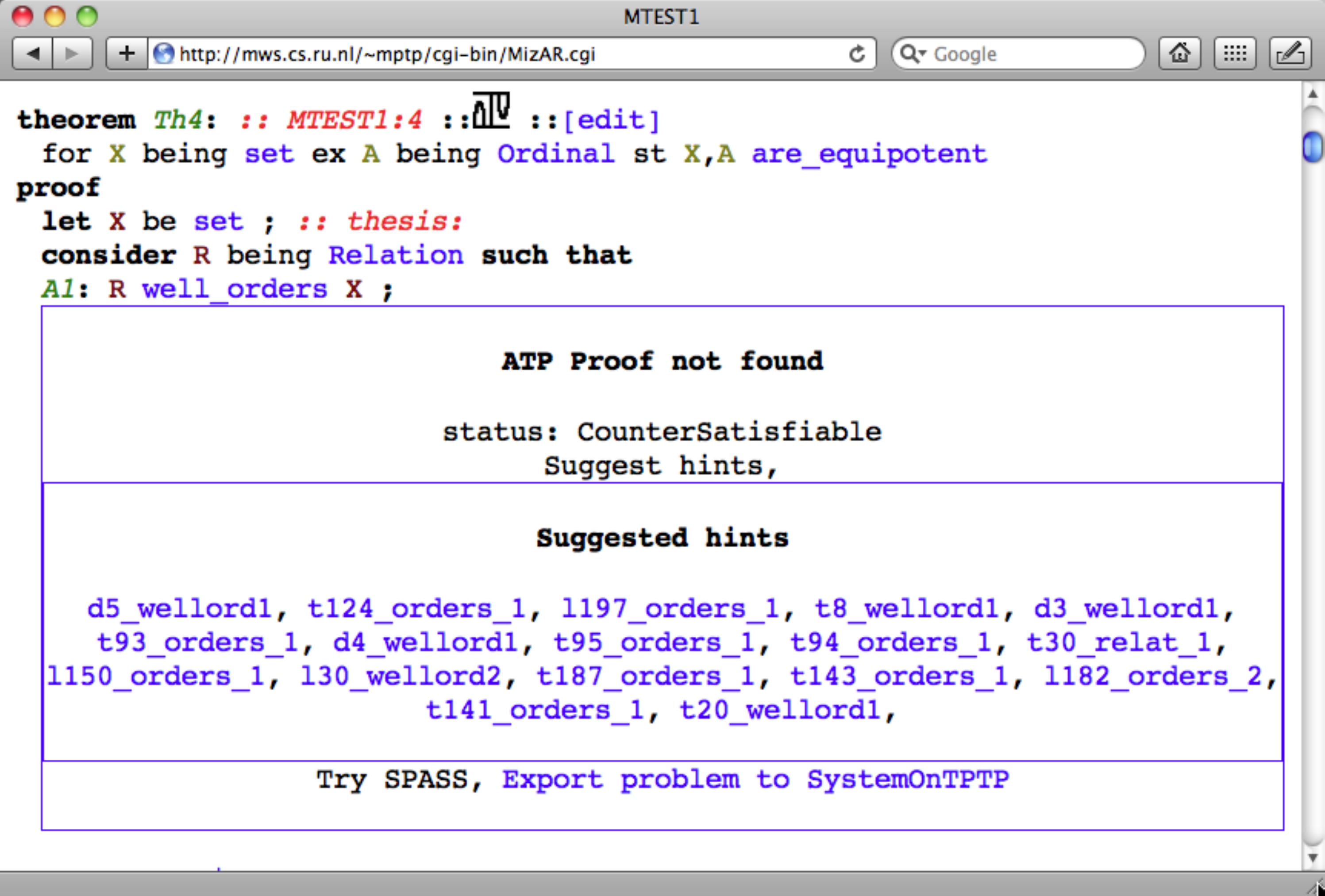}
  \caption{ATP explanation box offering hints}
  \label{Hints}
\end{center}
\end{figure}

If none of the ATP systems can find a proof for an ATP problem (corresponding
to a Mizar inference), either because the ATP system timed out or found
that the ATP problem is countersatisfiable (as in Figure~\ref{SPASS}, then 
typically some more assumptions (Mizar references) have to be added to 
the TPTP problem.
The explanation box can provide hints for proving the Mizar proposition.
This is done using the ``Suggest hints'' link that is put into the box
when EP fails to find a proof (see Figure~\ref{SPASS}).
The ``Suggest hints'' button is linked to a Bayesian advisor that has
been trained on the whole MML (i.e., on all of the proofs in it).
(See~\cite{US+08} for the details of how the machine
learning is organized in the context of a large deductive repository like MML.
Other axiom selection systems could be used in a similar way.)
The trained advisor runs as a daemon on the web server, and receives
queries initiated by clicking on the ``Suggest hints'' button.
This service is very fast, and the hints are usually provided in milliseconds.
They are HTMLized and inserted (by AJAX calls) into the explanation box, as
shown in Figure~\ref{Hints}.

\section{Future Work and Conclusions}
\label{Future}

This paper has introduced the {\sf MizAR} web service that allows Mizar users
to use automated reasoning tools on their Mizar articles.
{\sf MizAR} is to some degree based on and similar to the {\sf SystemOnTPTP} 
service for solving first-order ATP problems.
The main differences to {\sf SystemOnTPTP} are the use of the Mizar language
that is oriented towards human mathematicians (rather than the pure
first-order logic used in {\sf SystemOnTPTP}), and setting the service in
the context of the large Mizar Mathematical Library of previous theorems,
definitions, and proofs.

There are obvious differences to those systems, given by the large-theory
setting in which Mizar formalization is typically done.
There are several use-cases described above, ranging from using HTMLization
to disambiguate complicated Mizar syntax, usage of ATP systems to explain
Mizar inferences, provide new proofs, and find counterexamples, to using
additional AI-based services for proof advice, like the proof advisor
trained on the whole MML.

There are many directions for future work in this setting, some of them
mentioned above.
The service already is available by an interactive call from the Emacs
interface.
However, Emacs simply generates the request from the current Mizar buffer,
and lets the user see the response (and all the associated functionalities)
in a browser.
Obviously, the ATP and proof advising functionalities could be made
completely separate from the HTML presentation, and sent directly to the
Emacs session.

As mentioned above, the static MML is now present on the server in both
HTML and MPTP format (and obviously in raw text and in
the Mizar internal format), but not directly editable by the users.
Giving the user the ability to edit the supporting MML forms leads in the 
direction of formal mathematical wikis, with all the interesting persistence, 
versioning, linking, user-authentication, and dependency problems to solve.
There is an experimental ikiwiki-based prototype available for Mizar and
the MML, which solves some of the persistence and user-authentication problems,
and that is likely to be merged with the services presented here.
Hopefully this will form a rich wiki for formal mathematics, with a large
number of services providing interesting additional functionalities to people
interested in formal mathematics.

There is also a large amount of work that can be done on making the system
nicer and more responsive, e.g., the parallelization of the MPTP processing
is yet to be done, better machine learning and hint suggestion methods can
be used and linked to the ATP services, and presentations in formats other
than HTML (e.g., TeX and PDF are also used for Mizar) would be nice
to include.

\bibliographystyle{plain}
\bibliography{Bibliography}
\end{document}